\documentclass[a4paper,11pt]{article}

\usepackage[utf8]{inputenc}
\usepackage[british]{babel}
\usepackage{lmodern}
\usepackage[T1]{fontenc}
\usepackage[top=2.5cm, bottom=2.5cm, left=2cm, right=2.cm]{geometry}
\usepackage{amsmath}
\usepackage{ amssymb }

\usepackage{amsmath,amsfonts,bm}

\def\eqref#1{equation~\ref{#1}}

\def\1{\bm{1}}

\def\mI{{\bm{I}}}

\DeclareMathAlphabet{\mathsfit}{\encodingdefault}{\sfdefault}{m}{sl}
\SetMathAlphabet{\mathsfit}{bold}{\encodingdefault}{\sfdefault}{bx}{n}

\usepackage{hyperref}
\usepackage{url}
\usepackage{graphicx}

\usepackage{microtype}

\usepackage[utf8]{inputenc} 
\usepackage[T1]{fontenc}    
\usepackage{url}            
\usepackage{booktabs}       
\usepackage{amsfonts}       
\usepackage{nicefrac}       
\usepackage{microtype}      
\usepackage{xcolor}         
\usepackage[capitalize]{cleveref}

\def\be{\begin{equation}}
\def\ee{\end{equation}}
\def\bes{\begin{subequations}}
\def\ees{\end{subequations}}
\def\bea{\begin{eqnarray}}
\def\eea{\end{eqnarray}}
\def\bry{\begin{array}}
\def\ery{\end{array}}
\def\bit{\begin{itemize}}
\def\eit{\end{itemize}}
\def\ben{\begin{enumerate}}
\def\een{\end{enumerate}}

\def\tr{\textrm{Tr}}

\def\({\left(}
\def\){\right)}

\newcommand{\alonb}[1]{}

\newcommand{\loss}{\mathcal{L}}
\newcommand{\gradtheta}{\nabla_{\theta}}

\newcommand{\norm}[1]{\| #1 \|}
\newcommand{\dout}{d_\mathrm{out}}
\newcommand{\din}{d_\mathrm{in}}

\newcommand{\training}{\mathrm{tr}}
\newcommand{\gen}{\mathrm{gen}}

\newcommand{\mcL}{\mathcal{L}}

\title{\textbf{Grokking in Linear Estimators -- A Solvable Model that Groks without Understanding}}

\author{
\textbf{Noam~Levi, Alon~Beck \& Yohai~Bar~Sinai} \\
Raymond and Beverly Sackler School of Physics and Astronomy\\
Tel-Aviv University\\
Tel-Aviv 69978, Israel \\
\texttt{\{noam@,alonback@tauex,ybs@\}.tau.ac.il} \\
}

\begin{document}
\maketitle

\begin{abstract}
    Grokking is the intriguing phenomenon where a model learns to generalize long after it has fit the training data. 
    We show both analytically and numerically that grokking can surprisingly occur in linear networks performing linear tasks in a simple teacher-student setup with Gaussian inputs. 
    In this setting, the full training dynamics is derived in terms of the training and generalization data covariance matrix. 
    We present exact predictions on how the grokking time depends on input and output dimensionality, train sample size, regularization, and network initialization.
    We demonstrate that the sharp increase in generalization accuracy may not imply a transition from "memorization" to "understanding", but can simply be an artifact of the accuracy measure. We provide empirical verification for our calculations, along with preliminary results indicating that some predictions also hold for deeper networks, with non-linear activations.

\end{abstract}
\section{Introduction}

Understanding the underlying correlations in complex datasets is the main challenge of statistical learning. Assuming that training and generalization data are drawn from a similar distribution, the discrepancy between training and generalization metrics quantifies how well a model extracts meaningful features from the training data, and what portion of its reasoning is based on idiosyncrasies in the training data. Traditionally, one would expect that once a neural network (NN) training  converges to a low loss value, the generalization error should either plateau, for good models, or deteriorate for models that overfit.

Surprisingly, \cite{power2022grokking} found that a shallow transformer trained on algorithmic datasets features drastically different dynamics. The network first overfits the training data, achieving low and stable training loss with high generalization error for an extended period, then suddenly and rapidly transitions to a perfect generalization phase.

This counter-intuitive phenomenon, dubbed \emph{grokking}, has recently garnered much attention and many underlying mechanisms have been proposed as possible explanations. These include the difficulty of representation learning~\cite{liu2022towards}, the scale of parameters at initialization~\cite{liu2023omnigrok}, spikes in loss ("slingshots") \cite{thilak2022slingshot}, random walks among optimal solutions~\cite{millidge2022}, and the simplicity of the generalising solution~\cite[Appendix E]{nanda2023progress}.

In this paper we take a different approach, leveraging the simplest possible models which still display grokking - linear estimators. Due to their simplicity, this class of models offers analytically tractable dynamics, allowing a derivation of exact predictions for grokking, and a clear interpretation which is corroborated empirically. Our main contributions are:

\begin{itemize}
    \item We solve analytically the gradient-flow training dynamics in a linear teacher-student ($T, S \in \mathbb{R}^{\din\times \dout}$) model performing MSE classification. In this setting, the training and generalization losses $\mathcal{L}_\training, \mathcal{L}_\gen$, are simply given by $||T-S||^2_{\Sigma}$, where the norm is defined with respect to the training/generalization Gram matrices, $\Sigma_{\mathrm{tr}}$ and $\Sigma_{\mathrm{gen}}$ respectively. These matrices can be modeled with classical Random Matrix Theory (RMT) techniques.
    \item Grokking in this setting does not imply any ``interesting''  generalization behavior, but rather the simple fact that the generalization loss decays slower than the training loss, because the gradients are set by the latter. The grokking time is mainly determined by a single parameter, the ratio between input dimension and number of training samples $\lambda=\din/N_\training$.
    \item Standard variations are included in the analysis:
    \begin{itemize}
     \item The effect of different weight initializations is to generate an artificial rescaling of the training and generalization losses, increasing the effective accuracy value required for saturation and therefore increasing grokking time.
    \item For small $\dout$, Grokking time increases with output dimension due to effectively slower dynamics. This happens up to a critical dimension after which the measure of accuracy becomes insensitive to the value of the loss, reducing the grokking time.
    \item $L_2$ regularization suppresses grokking in overparameterized networks as expected, while having a subtle effect on the grokking time in underparameterized settings. 
    \end{itemize}
    \item We further show semi-analytically that our results extend to architectures beyond shallow linear networks, including one hidden layer, with both linear and some nonlinear activations.
\end{itemize}
\section{Related work}
\paragraph{Grokking}  
Many works have attempted to explain the underlying mechanism responsible for grokking, since its discovery by \cite{power2022grokking}. Some works suggest "slingshots" \cite{thilak2022slingshot} or "oscillations" \cite{notsawo2023predicting} underlie grokking, but our explanation applies even without these dynamics.
Other works identify ingredients for grokking \cite{davies2023unifying, nanda2023progress}, analyze the trigonometric algorithms networks learn after grokking \cite{nanda2023progress, chughtai2023toy, merrill2023tale}, and show similar dynamics in sparse parity tasks \cite{merrill2023tale}. 
The addition of regularization has been shown to strongly affect grokking in certain scenarios \cite{power2022grokking,liu2023omnigrok}. This connection may be attributed to weight decay (WD), for instance, improving generalization~\cite{Krogh1991ASW}, though this property is not yet fully understood~\cite{zhangWang}. We incorporate WD in our setup and study its effects on grokking analytically, showing that it can either suppress or enhance grokking, depending on the number of network parameters and number of training samples. 

Key related works, most closely tied with our own, are \cite{liu2022towards, liu2023omnigrok} and \cite{zunko2022grokking}. \cite{liu2022towards} show perfect generalization on a non-modular addition task when enough data determines the structured representation. \cite{liu2023omnigrok} relate grokking to memorization dynamics. \cite{zunko2022grokking} analyze solvable models displaying grokking and relate results to latent-space structure formation. Our work employs a similar setup but derives grokking dynamics from a random matrix theory perspective relating dataset properties to the empirical covariance matrix.

\paragraph{Linear Estimators in High Dimensions}  

A growing body of work has focused on deriving exact solutions for linear estimators trained on Gaussian data, particularly in the context of random feature models. The dynamics are often described in the gradient flow limit, which we employ in this work. Building on statistical physics methods, \cite{crisanti2018dynamics} provided an analytical characterization of the dynamics of learning in linear neural networks under gradient descent. Their mean-field analysis precisely tracks the evolution of the training and generalization errors, similar to \cite{richards2021asymptotics}. More recently, \cite{bodin2021dynamics, bodin2022gradient} further studied the dynamics of generalization under gradient descent for piecewise linear networks and for the Gaussian covariate model, corroborating the presence of epoch-wise descent structures. In the context of least squares estimation and multiple layers, \cite{Loureiro_2022,goldt2020modelling} analyzed the gradient flow dynamics and long-time behavior of the training and generalization errors. The tools from random matrix theory and statistical mechanics employed in these analyses allow precise tracking of the generalization curve and transitions thereof, akin to \cite{dobriban2015highdimensional}. Our work adopts a similar theoretical framing to study the interplay between model capacity, overparameterization, and gradient flow optimization in determining generalization performance.

\section{Training dynamics in a linear teacher-student setup}
\label{sec:training_dynamics}

The majority of our results are derived for a simple student teacher model \cite{seung1992statistical}, where the inputs are identical independently distributed (iid) normal variables. We draw $N_\training$ training samples from a standard Gaussian distribution $\mathcal{N}(0, \mI_{\din \times \din })$, and the teacher model generates output labels. The student is trained to mimic the predictions of the teacher, which we take to be perfect.

The teacher and student models, which we denote by $T$ and $S$ respectively, share the same architecture. As we show below, Grokking can occur even for the simplest possible network function, which is a linear Perceptron with no biases, or in other words -- a simple linear transformation. The loss function is the standard MSE loss. 
Our analyses are done in the regime of large input dimension and large sample size, i.e., $\din,N_\training\to \infty$, where the ratio $\lambda\equiv \din/N_\training \in \mathbb{R}^+$ kept constant.

Following the construction presented in~\cite{liu2023omnigrok}, we can convert this regression problem into a classification task by setting a threshold $\epsilon>0$ and defining a sample to be correctly classified if the prediction error is less than $\epsilon$.
The student model is trained with the full batch Gradient Descent (GD) optimizer for $t$ steps with a learning rate $\eta$, which may also include a weight decay parameter $\gamma$. 
The training loss function is given by\alonb{ add $=\frac{1}{d_{\mathrm{out}}}\norm{D}_{\Sigma_{\mathrm{tr}}}$?}
\begin{align}
\label{eq:linear_loss}
    \mathcal{L}_\training 
    &= 
    \frac{1}{N_\training \dout}
    \sum_{i=1}^{N_\training }
    \norm{ 
    (S  - T)^T x_i
    }^2 
    =
    \frac{1}{\dout}
    \tr
    \left[
    D^T \Sigma_\training D 
    \right]
    ,
    & 
    D&\equiv S-T
    \ .
\end{align}
where $S,T \in \mathbb{R}^{\din\times \dout}$ are the student and teacher weight matrices, $\Sigma_\training \equiv \frac{1}{N_\training}  \sum_{i=1}^{N_\training} x_ix^T_i$ is the $d_{\mathrm{in}} \times d_{\mathrm{in}}$ empirical data covariance, or Gram matrix for the {\it training} set, and we define $D$ as the difference between the student and teacher matrices.  The elements of $T$ and $S$ are drawn at initialization from a normal distribution $S_0, T \sim \mathcal{N}(0,1/({2 \din }\dout))$. We do not include biases in the student or teacher weight matrices, as they have no effect on centrally distributed data.

Similarly, the generalization loss function is defined as its expectation value over the input distribution, which can be approximated by the empirical average over $N_\gen$ randomly sampled points\alonb{and here $=\frac{1}{d_{\mathrm{out}}}\norm{D}_{\Sigma_{\mathrm{gen}}}$?}
\begin{align}
    \mathcal{L}_\gen 
    &= \mathbb{E}_{x\sim\mathcal{N}}\left[
    \frac{1}{\dout}
    \norm{ 
    (S  - T)^T x
    }^2\right]
    =\frac{1}{\dout}
    \tr\left[ 
    D^T \Sigma_\gen D 
    \right]
    = \frac{1}{\dout}\norm{D}^2 
    \ .
    \label{eq:generalization_linear_loss}
\end{align}
Here $\Sigma_\gen$ is the covariance of the generalization distribution, which is the identity. Note that in practice the generalization loss is computed by a sample average over an independent set, which is not equal to the analytical expectation value. 
The gradient descent equations at training step $t$ are \alonb{What does the left equation contribute?}
\begin{align}
\label{eq:GD}
    \nabla_D \loss_\training&=\frac{2}{\dout}\Sigma_\training D\ ,
    &
    {D}_{t+1}
    =
     \left( \mI  - \frac{2\eta}{d_\mathrm{out}} \Sigma_\training  \right) D_t
     - \frac{\eta \gamma}{\dout} \left(D_t\ +T\right) ,
\end{align}
where $\gamma \in \mathbb{R}^+$ is the weight decay parameter, and $\mI \in \mathbb{R}^{\din \times \din}$ is the identity.

It is worthwhile to emphasize the difference between \cref{eq:linear_loss} and \cref{eq:generalization_linear_loss}, since the distinction between sample average and analytical expectation value is crucial to our analyses. In training, \cref{eq:linear_loss}, we compute the loss over a fixed dataset whose covariance, $\Sigma_\training$, is non trivial. The generalization loss is defined as the expectation value over the input distribution, which has a trivial covariance by assumption, $\Sigma_\gen = \mI$. Even if it is computed in practice by averaging over a finite sample with a non trivial covariance, it is independent of the training dynamics and the sample average will converge to the analytical expectation with the usual $\sqrt N$ scaling. This is \textit{not true} for the training loss, since the training dynamics will guide the network in a direction that minimizes the empirical loss with respect to the fixed covariance $\Sigma_\training$. This assertion is numerically verified below, as we compare the generalization loss, practically computed by sample averaging, to the analytical result of \cref{eq:generalization_linear_loss}.

\subsection{Warmup: the simplest model}
\label{sec:simplified}
\subsubsection{Train and generalization loss}
Before analyzing the dynamics of the general linear model, we start with a simpler setting which captures the most important aspects of the full solution. 
Concretely, here we set $\dout=1$, reducing $S,T\in \mathbb{R}^{\din}$ from matrices to vectors, and assume no weight decay $\gamma=0$.
\cref{eq:GD} can be solved in the gradient flow limit of continuous time, setting $\eta = \eta_0 dt $ and $dt \to 0$, resulting in
\begin{align}
\label{eq:gdflow_sol}
\dot{D}(t) = -2\eta_0 \Sigma_\training D(t) 
\quad\to\quad
    D(t) =  e^{-2\eta_0 \Sigma_\training t} D_0,
\end{align}
where $D_0$ is simply the difference between teacher and student vectors at initialization.
It follows that the empirical losses, calculated over a dataset functions admit closed form expressions as
\begin{align}
\label{eq:losses_general_sol}
    \mathcal{L}_\training
    =
    D_0^T e^{-4\eta_0 \Sigma_\training t} \Sigma_\training D_0,
    \qquad
    \mathcal{L}_\gen
    =
    D_0^T 
    e^{-4\eta_0 \Sigma_\training t} 
    D_0.
\end{align}

\begin{figure}
    \centering
    \includegraphics[width=1.\textwidth]{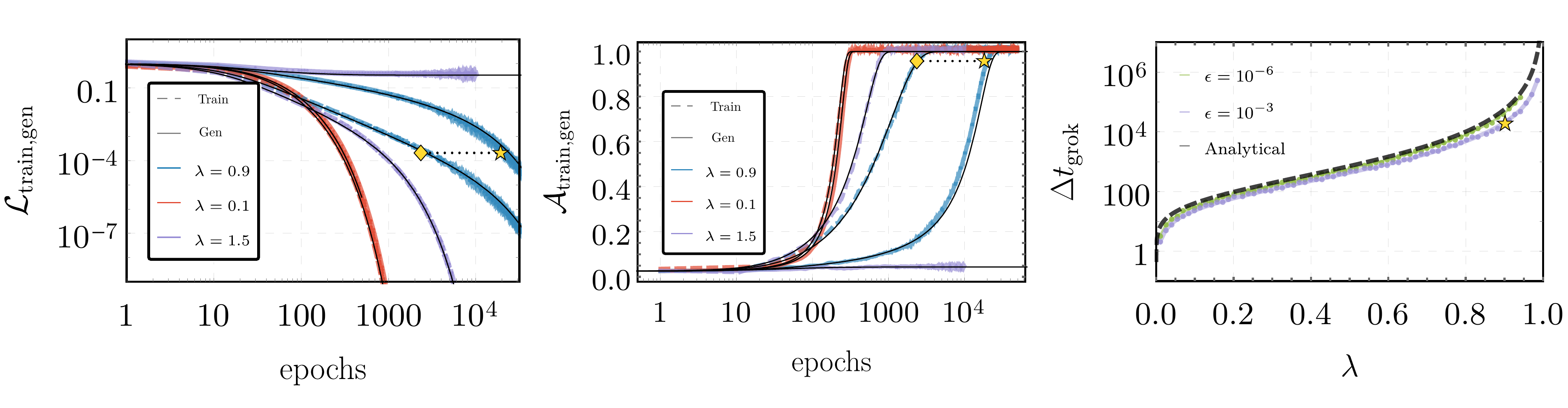} 
    \caption{
        Grokking as a function of $\lambda$.
    {\bf Left:} Empirical results for training (dashed) and generalization (solid) losses, for $\lambda=0.1,0.9,1.5$ (red, blue, violet) against analytical solutions (black).
    {\bf Center:} Similar comparison for the accuracy functions.
    {\bf Right:} The grokking time as a function of $\lambda$, for different values of the threshold parameter $\epsilon$.
    Different solid curves are numerical solutions for the expressions given in \cref{sec:simplified}, shown against the analytic solution in \cref{eq:t_grok_anal} (dashed black).
    In all three panels, {\bf diamonds/stars} indicate training/generalization accuracy convergence to 95\%.
    Training is done using GD with $\eta=\eta_0=0.01, \din=\!10^{3}, \dout=1, \epsilon = \! 10^{-3}$. 
      }
    \label{fig:toy_model}
\end{figure}

These expressions for the losses are exact. To proceed, we need to know the Gram matrix of the training dataset, which is the empirical covariance of a random sample of Gaussian variables. It is known that eigenvectors of $\Sigma_\training$ are uniformly distributed on the unit sphere in $\mathbb{R}^{\dout}$ and its eigenvalues, $\nu_i$, follow the Marchenko-Pastur (MP) distribution~\cite{VAMarcenko_1967},
\begin{align}
    \label{eq:MPD}
    p_\mathrm{MP}(\nu)d\nu
    =
 \left(1-\frac{1}{\lambda}\right)^+ \delta_{0} 
 +  \frac{\sqrt{(\lambda_+ - \nu)(\nu-\lambda_-)}}{2 \pi \lambda \nu}I_{\nu\in [\lambda_-, \lambda_+]}  d\nu ,
\end{align}
where $\delta_\nu$ is the Dirac mass at $\nu \in \mathbb{R}$, we define $x^+ = \max\{x,0\}$ for $x\in\mathbb{R}$, and $\lambda_{\pm}=(1\pm \sqrt{\lambda})^2$.\alonb{I think $\delta_{0}$ in the Eq. should be $\delta_{\nu}$ (or maybe less confusing to say $\delta_{\nu,0}$? And why define the $()^+$ notation if we use it only once?}

Since the directions of both $D$ and the  eigenvectors of $\Sigma_\training$ are uniformly distributed, we make the approximation that the projection of $D$ on all eigenvectors is the same, which transforms \cref{eq:losses_general_sol} to the simple form
\begin{align}
\label{eq:MP_loss_sums}
   \mathcal{L}_\training &\approx \frac{1}{\din}\sum_i e^{-4\eta_0 \nu_i  t}  \nu_i\ ,
    &
    \mathcal{L}_\gen &\approx \frac{1}{\din} \sum_i e^{-4 \eta_0\nu_i  t} \ .
\end{align}
It is seen that these sums are the empirical average over the function $e^{-4 \eta_0 \nu t} \nu$, if $\nu$ follows the MP distribution. This can be well approximated by their respective expectation values,
\begin{align}
\label{eq:train_loss_anal}
  \mathcal{L}_\training(\eta_0, \lambda, t) &\approx
  \mathbb{E}_{\nu\sim \mathrm{MP(\lambda)}}\left[ \nu e^{-4\eta_0  \nu t}\right]
  , &
  \mathcal{L}_\gen(\eta_0, \lambda, t) &\approx
  \mathbb{E}_{\nu\sim \mathrm{MP(\lambda)}}\left[ e^{-4\eta_0  \nu t}\right]
  .
\end{align}
The evolution of these loss functions are dictated by the MP distribution, which exhibits distinct behaviors for $\lambda \lessgtr 1$. For $\lambda<1$, the first term in~\cref{eq:MPD} vanishes, the distribution has no null eigenvalues and so $\mcL_\training, \mcL_\gen$ both are driven to $0$ at $t \to \infty$, implying that perfect generalization is always obtained eventually. On the other hand, for $\lambda>1$, \cref{eq:MPD} develops a number of zero eigenvalues, corresponding to flat directions in the training Gram matrix. In this case, while $\mcL_\training$ is driven to 0, since $\nu e^{-4\eta_0 \nu t}|_{\nu=0}=0$, the generalization loss $\mcL_\gen$ does not vanish, as $ e^{-4\eta_0 \nu t}|_{\nu=0}=1$ contributes a nonzero constant $1-\frac{1}{\lambda}$ to the loss, preventing perfect generalization. When $\dout=1$, the two regimes correspond directly to underparameterization ($\lambda<1$) and overparameterization ($\lambda>1$).

In \cref{fig:toy_model} we show that these analytical prediction are in excellent agreement with numerical experiments, with no fitting parameters, in both regimes. 

We also note that the expectation value of $\mathcal{L}_\training$ in \cref{eq:train_loss_anal} admits a closed form solution, 
\begin{equation}
    \mathcal{L}_\training=e^{-4 \eta_0 (\lambda +1) t} \, _0\tilde{F}_1\left(2;16 \eta_0^2 t^2 \lambda \right) ,
\end{equation}
where $_0\tilde{F}_1 \left(a;z\right)= {_0{F}_1} (a;z)\Gamma (a)$ is the regularized confluent hypergeometric function. We could not find a closed form expression for $\mathcal{L}_\gen$, but approximate expressions for the expectation value can be derived for the late time behavior, cf.~\cref{sec:app_grokking_time_difference}.

\subsubsection{Train and generalization accuracy}
Next, we describe the evolution of the training and generalization accuracy functions. 
As described above, in the construction of \cite{liu2023omnigrok} the accuracy $\mathcal{A}$ is defined as the (empirical) fraction of points whose prediction error is smaller than $\epsilon$, $\mathcal{A}
=
\frac{1}{N}
\sum_{i=1}^{N}
 \Theta( \epsilon -  (D^T(t) x_i)^2)$, where $\Theta$ is the Heaviside step function. We define $z=D^T x\in\mathbb{R}$, which is normally distributed with standard deviation $D^T \Sigma D=\loss$, where $\Sigma$  is the covariance of $x$ (that is, $\Sigma_\training$ for training and $\mI$ for generalization). Then, in the limit of large sample sizes the empirical averages converge to
\begin{align}
    \mathcal{A}_{\training/\gen}
    \to 2\Pr\left( |z| \leq  \sqrt{\epsilon} \right)
    =
    \mathrm{Erf}
\left(\sqrt{ \frac{\epsilon}{2 \mathcal{L}_{\training/\gen} }} \right) \ .
\label{eq:accuracy_as_function_of_loss}
\end{align}
The implication of this result is that the increase in accuracy in late stages of training can be simply mapped to the decrease of the loss below $\epsilon$. Writing the accuracy as an explicit function of the loss allows an exact calculation of the grokking time, and of whether grokking occurs at all. 

\subsubsection{Grokking time}
In this framework, grokking is simply the phenomenon in which $\loss_\training$ drops below $\epsilon$ before $\loss_\gen$ does. To understand exactly when these events happen, in \cref{sec:app_grokking_time_difference} we derive approximate results in the long time limit, $\eta_0 t \gg \sqrt{\lambda}$, showing that
\begin{align}
    \label{eq:approx_losses}
    \mathcal{L}_\training
    \simeq
    \frac{e^{-4 \eta_0  \left(1-\sqrt{\lambda }\right)^2 t}}{16 \sqrt{\pi } \lambda ^{3/4} (\eta_0  t)^{3/2}}
    , \quad
      \mathcal{L}_\gen
    \simeq  
    \mathcal{L}_\training
    \times
    \left(1-\sqrt{\lambda }\right)^{-2}.
\end{align}
We define grokking time as the time difference between the training and generalization accuracies reaching $\operatorname{Erf}(\sqrt{2})\approx95\%$, obtained when each loss satisfies $\mathcal{L}(t^*)= \epsilon/4$. In terms of the loss functions, we show in \cref{sec:app_grokking_time_difference} that solving for the difference between $t^*_\gen-t^*_\training$, and expanding the result in the limit of $\epsilon \ll 1$, one obtains an analytic expression for the grokking time difference 
\begin{align}
\label{eq:t_grok_anal}
    \Delta t_\mathrm{grok}
    =
    t^*_\gen - t^*_\training
    \simeq
    \tfrac{\log \left(\frac{1}{1-\sqrt{\lambda }}\right)}{2 \eta_0  \left(1-\sqrt{\lambda }\right)^2}.
\end{align}
\cref{eq:t_grok_anal} indicates that the maximal grokking time difference occurs near $\lambda \simeq 1$, where the grokking time diverges quadratically as $\Delta t_\mathrm{grok}(\lambda\to1)\sim\frac{1}{\eta _0 (\lambda -1)^2}\log \left(\frac{4}{(1-\lambda )^2}\right)$. On the other hand, it vanishes for $\lambda \simeq 0$, which means $N_\training \gg \din $ and $\Sigma_\training$ approaches the identity, as expected. These predictions are verified in \cref{fig:toy_model}(right).

\paragraph{Effects of Initialization and Label Noise:}
We briefly comment on the effect of choosing a different initialization for the student weights compared to the teacher weights, which is discussed in \cite{liu2023omnigrok}, as well as adding training label noise. In the first setup, rescaling the student weights $S \to \alpha S$ leads to a trivial rescaling of both the training and generalization loss functions as $\mathcal{L}\to \frac{1+\alpha^2}{2} \mathcal{L}$, which is tantamount to choosing a different threshold parameter $\epsilon \to \frac{2\epsilon}{1+\alpha^2}$, leaving the results unchanged.
In the case of training label noise $y\to y + \delta$ , where
$\delta \sim \mathcal{N}(0,\sigma_\delta^2)$, the student dynamics don't change, but the generalization loss function would receive a constant contribution, proportional to the noise variance $\sigma^2_\delta$. This contribution will simply imply that for small $\epsilon$, grokking to perfect generalization cannot occur, but rather just to some finite accuracy.

\subsubsection{Interpretation and intuition}
We conclude this section by summarizing and interpreting the analytical results for the simple 1-layer linear network with a scalar output and MSE loss. In this setting, the loss, which is an empirical average over a finite sample, is given by the norm of $D=S-T$, as measured by the metric defined by the covariance of the sample, $\loss = D^T\Sigma D$. While the generalization covariance is the identity by construction, the train covariance only approaches the identity in the limit $N_\training\gg \din$, and otherwise follows the Marchenko-Pastur distribution. 

The training gradients point to a direction that minimizes the training loss, which is $\norm{D}_{\Sigma_\training}$, and in the long time limit it vanishes exponentially. This must imply that the generalization loss, $\norm{D}_{\bm I}$, which is the norm of the same vector but calculated wirth respect to a different metric, also vanishes exponentially but somewhat slower. Since in this setting the accuracy is a function of the loss, grokking is identified as the difference between the times that the training and generalization losses fall below the fixed threshold $\epsilon/4$. We note that the fact that the accuracy is an explicit function of the loss is a useful peculiarity of this model. In more general settings it is not the case, though it is generally expected that low loss would imply high accuracy. 

However, it is noteworthy that nothing particularly interesting is happening at this threshold, and the loss dynamics are oblivious to its existence. In other words, grokking in this setting, as reported previously by \cite{liu2023omnigrok}, is an artifact of the definition of accuracy and does not represent a transition from ``memorization'' to ``understanding'', or any other qualitative increase in any generalization abilities of the network.

Our analysis can be easily extended to include other effects in more complicated scenarios, which we detail below. In all these generalizations the qualitative interpretation remains valid.

\subsection{The effect of $\dout$}
\label{sec:d_out_main}
We first extend our analysis to the case $\dout>1$. The algebra in this case is similar to what was shown in \cref{sec:simplified}. We provide the full derivation in \cref{sec:app_d_out_effect} and report the main results here.

The loss evolution follows the same functional form as \cref{eq:train_loss_anal}, with the replacement $\eta_0 \to \eta_0/\dout$. In addition, when $\dout>1$ the mapping between $\loss$ and $\mathcal{A}$, \cref{eq:accuracy_as_function_of_loss}, should be corrected since $\norm{z}^2=\norm{D^Tx}^2$ now follows a $\chi^2$ distribution and not a normal distribution, resulting in
\begin{align}
    \label{eq:d_out_metrics}
    \mathcal{L}_\mathrm{\training/gen}^{\dout\geq 1} 
    &= 
    \frac{1}{\dout} 
    \mathcal{L}_\mathrm{\training/gen}^{\dout= 1} 
    \left( \frac{\eta_0}{\dout }, \lambda, t \right)\ , &
\mathcal{A}_\mathrm{\training/gen} &= 
  1-\frac{\Gamma \left(\frac{\dout}{2},\frac{\dout \epsilon }{2 \mathcal{L}_\mathrm{\training/gen}}\right)}{\Gamma \left(\frac{\dout}{2}\right)},
\end{align}
where $\Gamma (a,z)=\int _z^{\infty }d t  e^{-t}t^{a-1}$ is the incomplete gamma function, and $\Gamma(z)=\int _0^{\infty }d t e^{-t} t^{z-1}$ is the gamma function. It is seen that $\mathcal{A}$ is still an explicit function of $\loss$, albeit somewhat more complicated.

\begin{figure}
    \centering
    \includegraphics[width=1\textwidth]{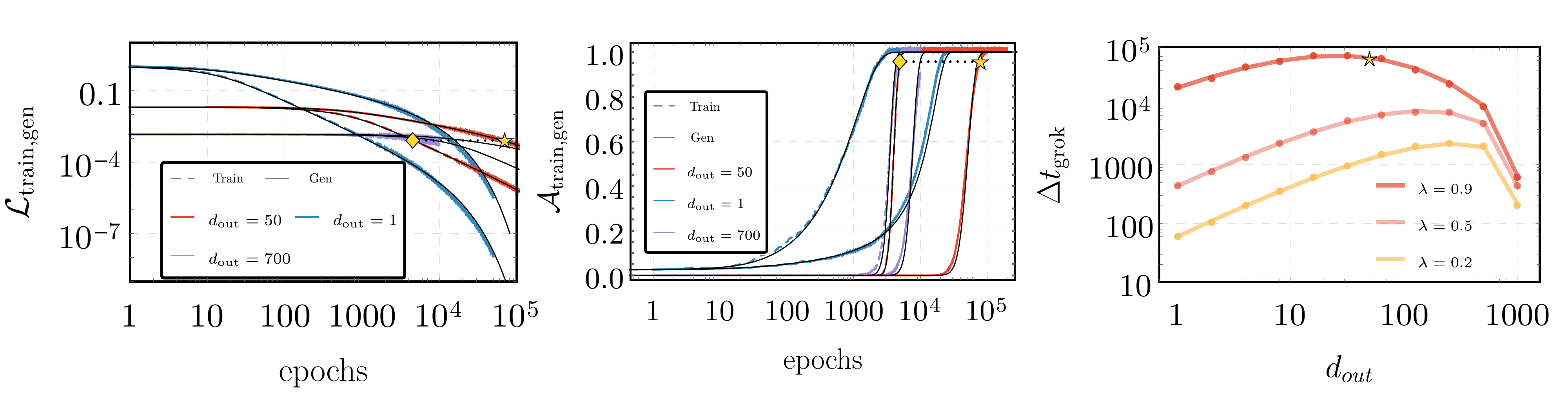}
    \caption{
    Effects of the output dimension $\dout>1$ on grokking.
    {\bf Left:} Empirical results for training (dashed) and generalization (solid) losses, for $\dout=1,50, 700$ (blue, red, violet) against analytical solutions (black), for $\lambda=0.9$.
    {\bf Center:} Similar comparison for the accuracy functions.
    {\bf Right:} The grokking time as a function of $\dout$, for different values of $\lambda$.
    Different solid curves are numerical solutions for the expressions given in \cref{sec:d_out_main}.
    In all three panels, {\bf diamonds/stars} indicate training/generalization accuracy convergence to 95\%, shown for $\dout^\mathrm{max}\simeq 50$, where the grokking time is maximal.
    Training is done using GD with $\eta=\eta_0=0.01, \din=\!10^{3}, \epsilon = \! 10^{-3}$.
      }
    \label{fig:effect_of_dout}
\end{figure}

The effects of $\dout>1$ can be read from \cref{eq:d_out_metrics}, and are twofold. Firstly, the accuracy rapidly approaches $1$ as the output dimension $\dout$ increases, for any value of $\loss$ and $\epsilon$. This implies that in the limit of $\dout\to \infty$, both training and generalization accuracies must be close to 100\% shortly after initialization and no grokking occurs. Secondly, the learning rate $\eta_0$ becomes effectively smaller as $\dout$ grows, implying that the overall time scale of convergence for both training and generalization accuracies increases, leading to a higher grokking time. 
These two competing effects, along with the monotonicity of the loss functions, give rise to a non-monotonic dependence of the grokking time on $\dout$, which attains a maximum at a specific value $\dout^\mathrm{max}$, as can be seen in \cref{fig:effect_of_dout}.

\subsection{The effect of weight decay}
We consider first the case of nonzero WD in the simpler case of $\dout=1$. Incorporating weight decay amounts to adding a regularization term at each gradient descent timestep, modifying \cref{eq:GD} to
\begin{align}
    \label{eq:GD_WD}
    D_{t+1}
    =
    D_t-2\eta\left(\Sigma_\training+\frac{1}{2}\gamma I\right)
    D_t-\eta\gamma T,
\end{align}
where $\gamma \in \mathbb{R}^+$ is the weight decay parameter. Comparing to \cref{eq:GD}, it is seen that this basically amounts to shifting the eigenvalues of $\Sigma$ by $\gamma$. The calculations are straightforward and detailed in \cref{sec:app_WD_derivation}, the result being that \cref{eq:train_loss_anal} is modified to read
\begin{align}
    \label{eq:WD_losses}
    \mathcal{L}_{\mathrm{\training/gen}}
    =
    \frac{1}{2}\mathbb{E}_{\nu\sim \mathrm{MP}(\lambda)}
    \left[
    \left(
        e^{-4\eta_0\left(\nu+\frac{1}{2}\gamma\right)t}
         +\left(\frac{e^{-2\eta_0\left(\nu+\frac{1}{2}\gamma\right)t}\nu+\frac{1}{2}\gamma}{\nu+\frac{1}{2}\gamma}\right)^{2}
        \right)
         q_{\training/ \gen}
        \right],
\end{align}
where $q_\training=\nu$ and $q_\gen=1$. Since $\gamma$ only affects the gradient but not the accuracy, the expression in \cref{eq:accuracy_as_function_of_loss} of $\mathcal{A}$ as a function of $\loss$, remains unchanged.

\begin{figure}
    \centering
    \includegraphics[width=1\textwidth]{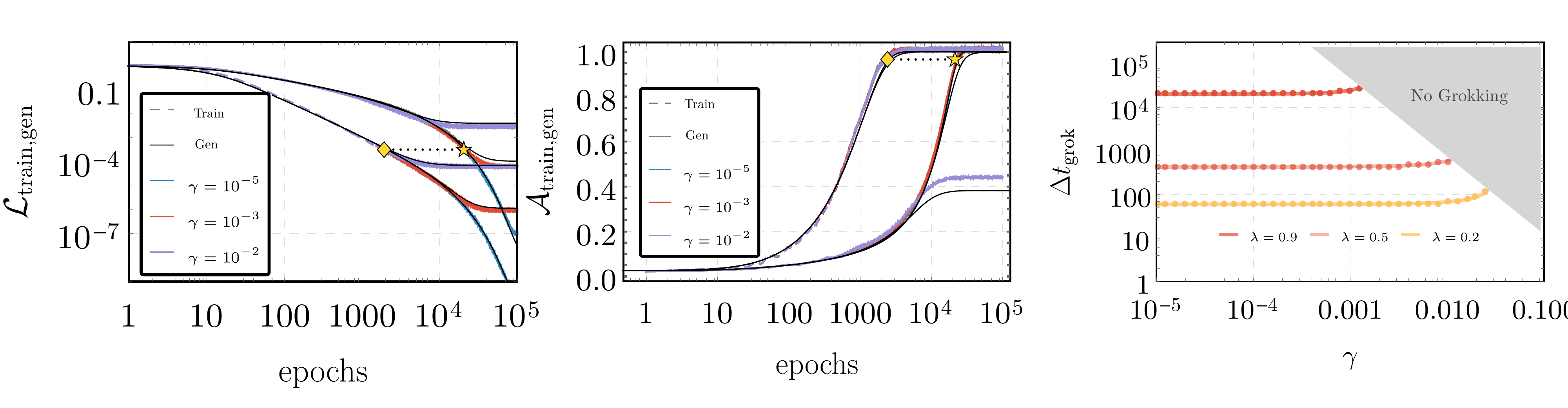} 
    \caption{
    Effects of weight decay ($\gamma$) on grokking.
    {\bf Left:} Empirical results for training (dashed) and generalization (solid) losses, for $\gamma=10^{-5},10^{-3}, 10^{-2}$ (blue, red, violet) against analytical solutions (black), for $\lambda=0.9$.
    {\bf Center:} Similar comparison for the accuracy functions.
    {\bf Right:} The grokking time as a function of $\gamma$, for different values of $\lambda$.
    Different solid curves are numerical solutions for the expressions given in \cref{sec:d_out_main}, while the shaded gray region corresponds to training/generalization saturation, without perfect generalization.
    In all three panels, {\bf diamonds/stars} indicate the point where accuracy reaches 95\%.
    Training is done using GD with $\eta=\eta_0=0.01, \din=\!10^{3}, \dout=1, \epsilon = \! 10^{-3}$
      }
          \label{fig:effect_of_WD}
\end{figure}
\begin{figure}
          \includegraphics[width=1\textwidth]{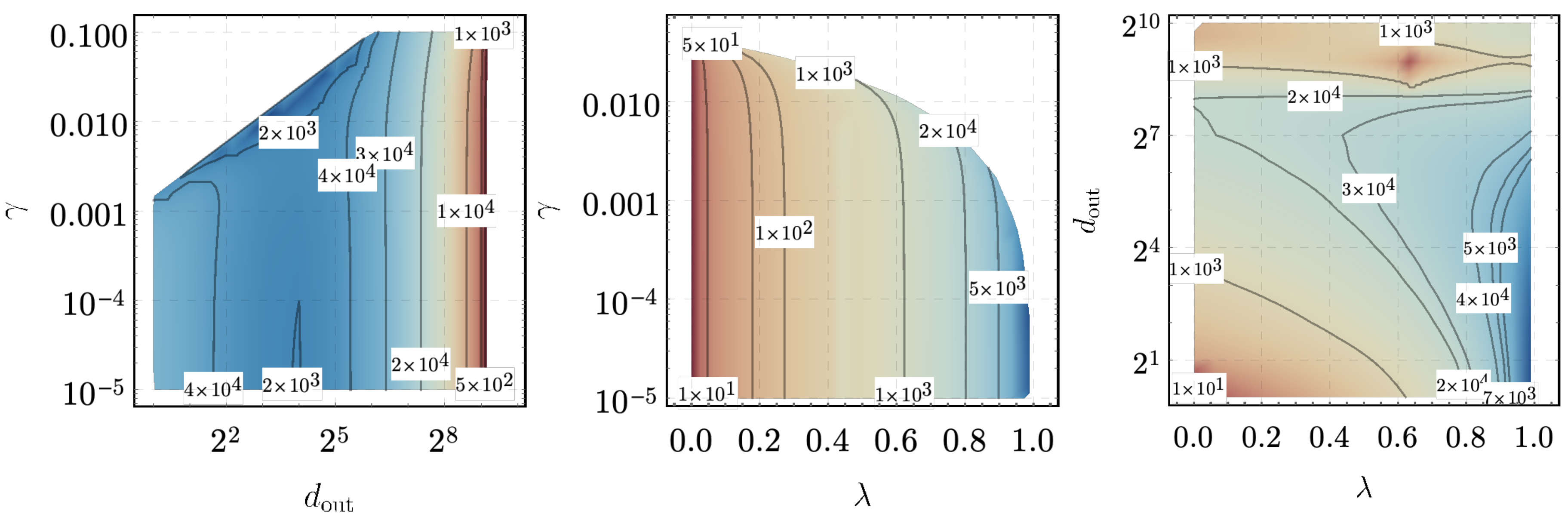}
\caption{
    Grokking time phase diagrams.
    {\bf Left:} 
    A contour plot of the grokking time difference as a function of $\gamma, \dout$. Shades of red indicate shorter grokking time, while blue tones indicate longer grokking time. White regions indicate no grokking, as generalization accuracy does not converge to $95\%$.
    {\bf Center} and {\bf Right:} Similar phase diagrams for the grokking time difference as a function of $\gamma, \lambda$ and $\dout, \lambda$, respectively.
    The results of all three plots are obtained by numerically finding the grokking time, using the definition $\mathcal{A}(t^*)=0.95$ and the analytic formulas quoted in the main text. The fixed parameters for these plots are $\eta_0=0.01, \epsilon = \! 10^{-3}$. }
    \label{fig:phase_diagrams}
\end{figure}

It is instructive to analyze \cref{eq:WD_losses} separately for the under and overparameterized regimes. When $\lambda<1$, the MP distribution has no null eigenvalues, and the losses begin by decaying exponentially. We can study the grokking behavior by examining the late time limit, i.e. $t\to\infty$, in which the exponential terms decay, and approximating for small $\gamma \ll 1$, we obtain the asymptotic expressions
\begin{align}
\label{eq:underparam_WD}
    \mcL_\training \simeq
    \frac{\gamma ^2}{4(1- \lambda )},
    \qquad
    \mcL_\gen
    \simeq
    \frac{\gamma ^2}{4 (1-\lambda )^3}
    , \qquad
    \Delta t_\mathrm{grok}
    \simeq
    \tfrac{\log 
    \left(
    1 + \sqrt{\lambda }
    \right)}{2 \eta _0 \left(1-\sqrt{\lambda }\right)^2}.
\end{align}
This result means that the generalization loss has a higher asymptotic value than the training loss. Thus, there is a value of $\epsilon$ below which perfect generalization cannot be obtained. For $\epsilon$ above this threshold WD has no effect, and below it the grokking time decreases as given by \cref{eq:underparam_WD}.

In the overparameterized regime, where $\lambda>1$, the MP distribution necessarily contains vanishing eigenvalues, which, as shown in \cref{fig:toy_model}, cause the generalization loss to plateau. Introducing weight decay changes this picture somewhat, causing the null eigenvalues to be shifted by a factor of $\gamma/2$ and ensuring that better generalization performance is reached. Still, the late time behavior is the same as \cref{eq:underparam_WD}, following the same arguments as discussed above. We note that in this case, the relevant timescale of the generalization loss is determined by $1/\gamma$, leading to suppressing grokking, as noted by \cite{liu2023omnigrok}.

The grokking time behaviors for various values of $\gamma$ are clearly demonstrated in \cref{fig:effect_of_WD}.

In \cref{fig:phase_diagrams}, we summarize our results by combining all the separate effects, showing two dimensional slices of the grokking phase diagram, which depends on $\lambda,\dout$ and $\gamma$, mirroring each separate effect.

\section{Generalizations}
\subsection{2-layer networks}
\label{sec:2layer}

Our analysis can be generalized to multi-layer models. Here, we consider the addition of a single hidden layer, where the teacher network function is\alonb{TODO: Check all of the $din \times dout$ and etc.} $f(x)= (T^{(1)})^T\sigma(( T^{(0)}) ^T  x)$, where $T^{(0)} \in \mathbb{R}^{\din \times d_h}$, $T^{(1)} \in \mathbb{R}^{ d_h\times \dout}$, $\sigma$ is an entry-wise activation function and $d_h$ is the width of the hidden layer. Similarly, the student network is defined by two matrices $S^{(0)}, S^{(1)}$. The empirical training loss reads
\begin{align}
\label{eq:2_layer_loss}
    \mathcal{L}_\training 
    &= 
    \frac{1}{N_\training \dout}
    \sum_{i=1}^{N_\training }
    \left( 
    (S^{(1)})^T\sigma((S^{(0)})^T  x_i)
    - (T^{(1)})^T \sigma ((T^{(0)})^T x_i)
    \right)^2\ .
\end{align}

In this setup, the weights are drawn at initialization from normal distributions~$S^{(0)}_0, T^{(0)} \sim \mathcal{N}(0,1/({2 \din }d_{h}))$ and~$S_0^{(1)}, T^{(1)} \sim \mathcal{N}(0,1/({2 \dout }d_{h}))$. 

\begin{figure}
    \centering
\includegraphics[width=.8\textwidth]{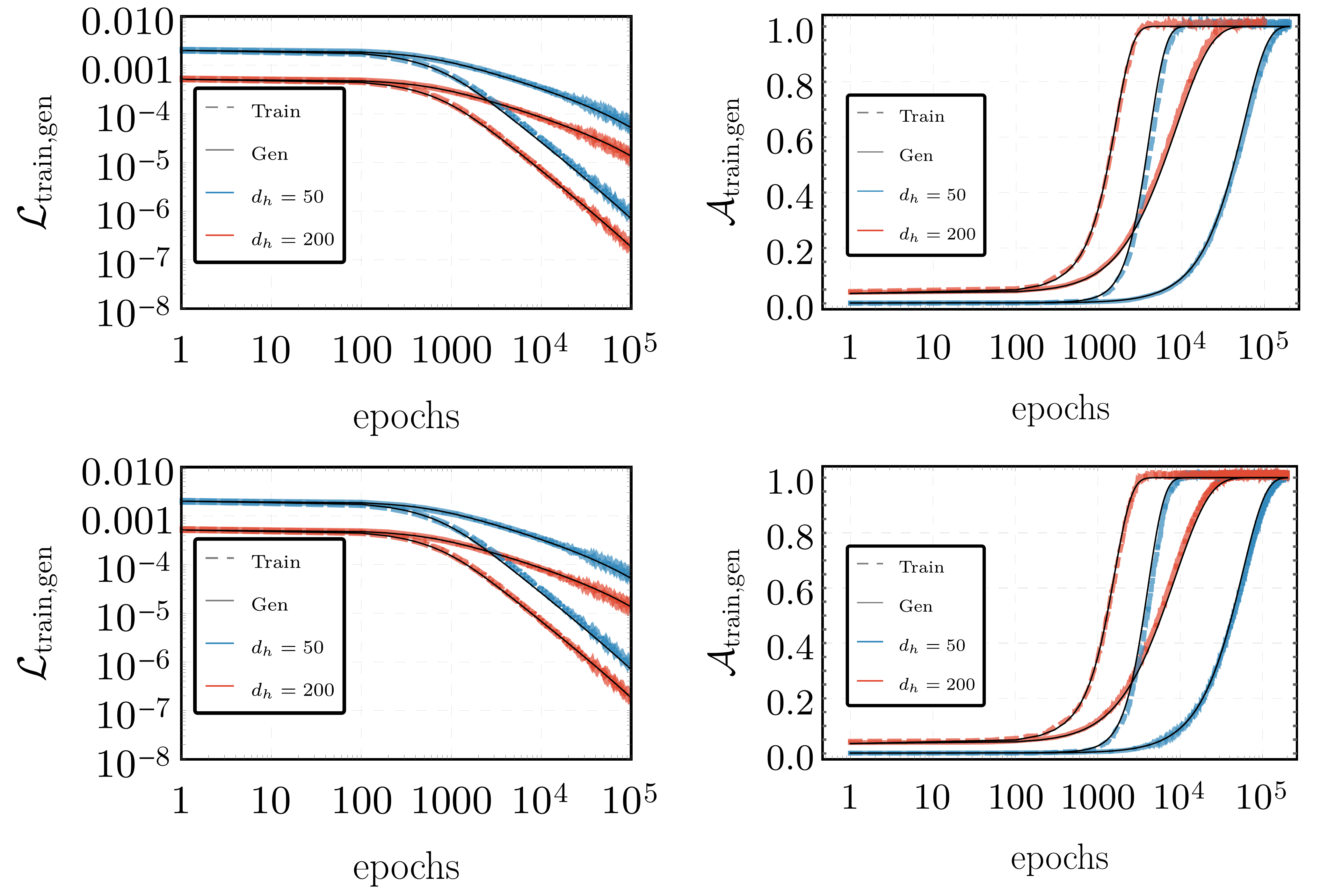}
    \caption{
    2-Layer network and nonlinearities. 
    {\bf Top row:} Empirical results for training (dashed) and generalization (solid) losses/accuracies ({\it left/right}), for a two layer MLP (1000-$d_h$-5) with linear activations and $d_h=50,200$ (blue, red), against analytical solutions (black).
    {\bf Bottom row:} Similar results, for a two layer MLP (1000-$d_h$-5) with $\tanh$ activations in the hidden layer.
    In both cases, training is done using full batch gradient descent with $\eta=\eta_0=0.01, \din=1000, \dout=5, \epsilon = 10^{-4}$. 
      }
    \label{fig:2_layers}
\end{figure}

As a solvable model, we consider first the case of linear activation, $\sigma(z)=z$, i.e., a two layer linear network. In this case we can define $T= T^{(0)} T^{(1)}  \in \mathbb{R}^{ \din \times \dout }$ as we did in the previous sections, since the teacher weights are not updated dynamically.
Similar to \cref{eq:linear_loss,eq:generalization_linear_loss}, under the definition $D_t= S_t^{(0)} S_t^{(1)} - T$, we show in \cref{sec:app_2layer} that the gradient flow equations for the system are
\begin{align}
\label{eq:GF_2layer}
    \dot{D}_t
    =
    -
    2\eta_0 \frac{h}{d_\mathrm{out}^2}  \Sigma_\training  D 
    ,
    \qquad
    \dot{h}_t
    =
    -8\eta_0 (T+D)^T \Sigma_\training D.
\end{align}
Here, $h=Tr[H]/2=\|S^{(0)}\|^2/2 + \|S^{(1)}\|^2/2$, where $H=\gradtheta^T \gradtheta \mathcal{L}_\training$ is the Hessian matrix and $\theta\equiv\{S^{(0)},S^{(1)}\}$.
Although \cref{eq:GF_2layer} describes a set of coupled equations, we note that the solution for $h_t$ can be simplified when considering the limit of small $\eta_0 \ll1$, as we may ignore the time evolution and consider the trace (or kernel) as fixed to its initialization value, which is $h_0\simeq 1/2$ for $d_h \gg d_\mathrm{out}$. In that case the loss solutions are a simple modification to the ones given in the previous sections, with the replacement $\eta_0 \to \eta_0 /(2 d_\mathrm{out}^2)$.
Subsequently, the training/generalization performance metrics are 
\begin{align}
\label{eq:non_linear_metrics}
  \mathcal{L}_\mathrm{\training/gen}^\mathrm{2-layer} 
  &= 
  \|D_0 \|^2
  \mathcal{L}_\mathrm{\training/gen}^\mathrm{1-layer}
  \left(
  \frac{\eta_0 }{2 \dout^2}, \lambda, t
  \right),
  \quad
    \mathcal{A}_\mathrm{\training/gen} = 
  1-\frac{\Gamma \left(\frac{\dout}{2},\frac{\dout \epsilon }{2 \mathcal{L}_\mathrm{\training/gen}}\right)}{\Gamma \left(\frac{\dout}{2}\right)} .
\end{align}
We note that this setup can be generically classified within the overparameterized regime, provided that $d_h \gg 1$, regardless of $\dout$ and for any $\lambda$. In this sense, all of the results previously derived for $\lambda<1$ hold, and grokking occurs as discussed in the previous sections. We experimentally verify that \cref{eq:non_linear_metrics} correctly predicts the performance metrics and their dynamics in \cref{fig:2_layers} (top row).

\subsection{Non-linear Activations}
\label{sec:nonlinear}
The final extension of our work is to consider the network in \cref{sec:2layer}, but choosing nonlinear activation functions for the hidden layer. In the limit of large $d_h \gg1$, we expect the network to begin to linearize, eventually converging to the Neural Tangent Kernel (NTK) regime \cite{lee2019wide}. In this regime, the results in~\cref{sec:2layer} should hold, up to a redefinition of the kernel which depends on the nonlinearity.

In \cref{fig:2_layers} (bottom row), we show that the dynamics of a 2 layer MLP (1000-200-5) with $\tanh$ activations is well approximated by \cref{eq:GF_2layer,eq:non_linear_metrics}, empirically verifying that our predictions hold beyond the linear regimes, in some cases.

\section{Discussion}
\label{sec:discussion}

We have shown that grokking can occur in simple linear teacher-student settings, and provided explicit analytical solutions for the training and generalization loss and accuracy dynamics during training. The predictions, which strictly apply in the gradient-flow limit and for large sample sizes, were corroborated against numerical experiments and provide an excellent description of the dynamics. In addition, preliminary evidence shows that some of the results are applicable also beyond the linear 1-layer setting, and are also pertinent for deeper networks and in the presence of non linearity.

Qualitatively, for linear networks with MSE loss, the training and generalization losses are given by the squared norm of the difference between the student and teacher weights, calculated with respect to the metric defined by the respective covariance matrices. Training reduces both norms, and grokking in this context simply reflects the fact that the generalization loss lags behind the training loss. However, no qualitative change in the behavior occurs at grokking, and consequently in this setting grokking does not imply any transition between memorization and understanding.

It would be interesting to go beyond the gradient flow limit, and study multilayer networks in the large learning rate regime, combining catapult/edge of stability dynamics with grokking analysis. 
Additionally, studying the effect of different optimizers on grokking could provide insights into how algorithmic choices influence the memorization to generalization transition. Furthermore, extending the grokking analysis to non-gaussian data or correlated inputs could reveal how data structure and correlations affect understanding versus memorization. Finally, in ongoing work, we consider more realistic accuracy measures such as softmax, or cross-entropy instead of mean squared error and connect these theoretical studies to practical deep learning settings. Overall, understanding grokking by building upon the insights provided by the linear estimator analysis could lead to a deeper understanding of how artificial neural networks balance fitting the training data with generalizing to new examples.

\section{Acknowledgements}
\label{sec:ack}
We thank Nadav Cohen for fruitful discussions. YBS was supported by research grant ISF 1907/22 and Google Gift grant. NL would like to thank the Milner Foundation for the award of a Milner Fellowship. This work was initiated in part at Aspen Center for Physics, which is supported by National Science Foundation grant PHY-2210452.

\bibliographystyle{plain}

\bibliography{grok}

\newpage

\appendix

\section{Experimental Details}
\label{sec:app_experimental_details}

In all of our experiments, we employ a teacher-student model with shared architecture for both teacher and student. The training data consists of a fixed number of training samples quoted in the main text for each experiment, drawn from a normal distribution~$\mathcal{N}(0,\mI)$. All experiments are done on MLPs using MSE loss with the default definitions employed by PyTorch. The exact details of each MLP depend on the setup and are quoted in the main text. We train with full batch gradient descent, in all instances. We depart from the default weight initialization of PyTorch, using $w \sim \mathcal{N}(0, 1/(2 d_{l-1} d_l)$ for each layer, where $d_{l-1}$ is in the input dimension coming from the previous layer and $d_l$ is the output dimension of the current layer.

\section{Derivation of the grokking time difference}
\label{sec:app_grokking_time_difference}

\begin{figure}
    \centering
\includegraphics[width=.65\textwidth]{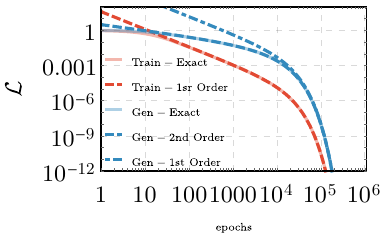}
    \caption{
        \alonb{Axis are too big?}Exact training and generalization losses against approximate solutions at late times. In {\bf pink/light blue}, we show the solutions of \cref{eq:train_loss_anal}. In {\bf dashed red} is \cref{eq:app_ltrain_aprox}, in {\bf dashed blue}, we show \cref{eq:app_lgen_aprox}, while {\bf dotted-dashed blue} is the solution given in the main text, \cref{eq:approx_losses}. Clearly, the asymptotic behavior matches the exact solutions. Here, $\eta_0=0.01, \lambda = 0.9, \dout=1.$
      }
    \label{fig:app_loss_aprox}
\end{figure}

Here, we provide the full derivation for the grokking time difference presented in \cref{eq:t_grok_anal}.
Our starting point is the exact solution for the training loss in $\dout=1$ case for $\lambda<1$, given by
\begin{align}
    \mathcal{L}_\training=e^{-4 \eta_0 (\lambda +1) t} \, _0\tilde{F}_1\left(2;16 \eta_0^2 t^2 \lambda \right),
\end{align}
where $_0\tilde{F}_1 \left(a;z\right)= {_0{F}_1} (a;z)\Gamma (a)$ is the regularized confluent hypergeometric function.
We also note the relation 
\begin{align}
\label{eq:app_lgen_ltrain}
    \frac{d \mcL_\gen}{dt }
    =
    -4\eta_0 \mcL_\training,
\end{align}
which we will use to relate training and generalization loss functions.
Since we are interested in the late time behavior, where grokking occurs, we expand the training loss for $\eta_0 t \gg \sqrt{\lambda}$, which is given at leading order by
\begin{align}
\label{eq:app_ltrain_aprox}
    \mathcal{L}_\training
    \simeq
    \frac{e^{-4 \eta_0  \left(1-\sqrt{\lambda }\right)^2 t}}{16 \sqrt{\pi } \lambda ^{3/4} (\eta_0  t)^{3/2}}.
\end{align}
Plugging in the result of \cref{eq:app_ltrain_aprox} into \cref{eq:app_lgen_ltrain} and integrating over time, we find the expression for the generalization loss at late times is given by
\begin{align}
\label{eq:app_lgen_aprox}
  \mcL_\gen
    \simeq
\frac{\sqrt{\eta_0  t} e^{-4 \eta_0  \left(1-\sqrt{\lambda }\right)^2 t}}{2 \sqrt{\pi } \eta_0  t \lambda ^{3/4} }-\frac{\left(1-\sqrt{\lambda }\right) \Gamma \left(\frac{1}{2},4  \eta_0 t  \left(1-\sqrt{\lambda }\right)^2\right)}{\sqrt{\pi } \lambda ^{3/4}},
\end{align}
where $\Gamma (a,z)=\int _z^{\infty }d t  e^{-t}t^{a-1}$ is the incomplete gamma function. Expanding the result further for late times, we arrive at the result quoted in \cref{eq:approx_losses}.
In \cref{fig:app_loss_aprox}, we show the approximate late time solutions against the exact solutions. The approximations hold quite well even at somewhat early times, and become increasingly more accurate for later epochs.

With the loss functions at hand, we turn to the grokking time itself. As described in the main text, we define the grokking time as the time difference between the training and generalization accuracies reaching $\operatorname{Erf}(\sqrt{2})\approx95\%$, obtained when each loss satisfies $\mathcal{L}(t^*)= \epsilon/4$.
Solving this equation for each loss separately, in the late time limit, gives the following expressions for the training and generalization times
\begin{align}
    t^*_\training
    &\simeq
    \frac{3 }{8 \eta _0 \left(1-\sqrt{\lambda }\right)^2}
    \mathcal{W}\left(\frac{2\ 2^{2/3} \sqrt[3]{\frac{\lambda ^{3/2}}{\pi }+\frac{1}{\pi  \lambda ^{3/2}}-\frac{6 \lambda }{\pi }+\frac{15 \sqrt{\lambda }}{\pi }-\frac{6}{\pi  \lambda }+\frac{15}{\pi  \sqrt{\lambda }}-\frac{20}{\pi }}}{3 \epsilon^{2/3}}\right),
    \\
    t^*_\gen 
    &\simeq
    \frac{3 }{8 \eta  \left(1-\sqrt{\lambda }\right)^2}
    \mathcal{W}\left(\frac{ 2^{5/3} \sqrt[3]{\frac{1}{\pi  \lambda ^{3/2}}+\frac{1}{\pi  \sqrt{\lambda }}-\frac{2}{\pi  \lambda }}}{3 \epsilon ^{2/3}}\right),
\end{align}
where $\mathcal{W}(z)$ is the Lambert W function, which solves the equation $\mathcal{W} e^\mathcal{W}=z$, also known as the product-log function. As the argument of both training and generalization times are large, we can expand the Lambert function to leading order in $z$ as $\mathcal{W}(z)\simeq \log(z)$. Taking the difference $\Delta t_\mathrm{grok}= t^*_\gen - t^*_\training$ and expanding to leading order in $\epsilon \ll1$, we obtain the final expression
\begin{align}
\label{eq:t_grok_anal_corrected}
    \Delta t_\mathrm{grok}
    =
    t^*_\gen - t^*_\training
    \simeq
    \frac{\log \left(\frac{1}{1-\sqrt{\lambda }}\right)}{2 \eta_0  \left(1-\sqrt{\lambda }\right)^2}
    +
    \frac{3 }{8 \eta  \left(1-\sqrt{\lambda }\right)^2}
    \log \left(
    1+
    \frac{\log \left(\left(1-\sqrt{\lambda }\right)^{4/3}\right)}{\log \left(\frac{2 \left(2-2 \sqrt{\lambda }\right)^{2/3}}{3 \sqrt[3]{\pi } \sqrt{\lambda } {\epsilon}^{2/3}}\right)}
    \right),
\end{align}
where the second term goes to zero as $\epsilon \to 0$, quoted in the main text as \cref{eq:t_grok_anal}.

\section{Derivation for $\dout>1$}
\label{sec:app_d_out_effect}

Here, we provide additional details on the derivation of \cref{eq:d_out_metrics}.
The starting point is the training and generalization loss functions, given by
\begin{align}
    \mathcal{L}_\training
    =
    \frac{1}{\dout}
    \tr
    \left[
    D^T \Sigma_\training D 
    \right]
    ,
    \qquad
    \mathcal{L}_\gen 
    =\frac{1}{\dout}
    \tr\left[ 
    D^T \Sigma_\gen D 
    \right]
    = \frac{1}{\dout}\norm{D}^2 
    \ .
\end{align}
where $S,T \in \mathbb{R}^{\din\times \dout}$ are the student and teacher weight matrices, $\Sigma_\training \equiv \frac{1}{N_\training}  \sum_{i=1}^{N_\training} x_ix^T_i$ is the empirical data covariance, or Gram matrix for the {\it training} set, and we define $D\equiv S-T$, the difference between the student and teacher matrices.  $T$ and $S$ are drawn at initialization from normal distributions~$S_0, T \sim \mathcal{N}(0,1/({2 \din }\dout))$. We do not include biases in the student or teacher weight matrices, as they have no effect on centrally distributed data.
The gradient descent equations in this instance are simply
\begin{align}
\label{eq:app_dout_GD}
    {D}_{t+1}
    =
     \left( \mI  - \frac{2\eta}{d_\mathrm{out}} \Sigma_\training  \right) D_t ,
\end{align}
where the only difference between the $\dout=1$ case and the equation above is the rescaled learning rate $\eta \to \eta/\dout$ and the dimensions of $D_t$. Since the MP distribution is identical for each column of $D_t$, the results sum up and are identical to the $\dout=1$ case for the losses, apart from a factor of $1/\dout$ and the learning rate rescaling, leading to \cref{eq:d_out_metrics}.


\section{Loss calculations for Dynamics including Weight Decay }
\label{sec:app_WD_derivation}

Here, we provide the derivation for \cref{eq:WD_losses}.
We begin with the definitions of the loss function in the $\dout=1$ case
\begin{equation}
    \mathcal{L}_\training=D(t)^{T}\Sigma_{\training}D(t), 
\end{equation}
where $D(t)=S(t)-T$ is the difference between the student and the
teacher vectors, $\Sigma_{\training}=\frac{1}{N_{\training}}\sum_{i=1}^{N_{\training}}x_{i}x_{i}^{T}$
is the training covariance matrix, and $\gamma\ge0$ is the weight decay
parameter. 
Using the gradient descent equation in the gradient flow limit, $\frac{\partial D}{\partial t}=-\eta\nabla_{D}\mathcal{L}$,
we obtain from \cref{eq:GD} that
\begin{equation}
    \frac{\partial D}{\partial t}=-2\eta\left(\Sigma_{\training}+\frac{1}{2}\gamma I\right)D-\eta\gamma T.
\end{equation}
Multiplying by the integration factor $e^{2\eta\left(\Sigma_{\training}+\frac{1}{2}\gamma I\right)t}$
and taking the integral, we arrive at
\begin{equation}
    D(t)
        + \frac{1}{2}\gamma
        \left(\Sigma_{\training}+\frac{1}{2}\gamma I\right)^{-1}T=e^{-2\eta\left(\Sigma_\training+\frac{1}{2}\gamma I\right)t}\left[D(0)+\frac{1}{2}\gamma\left(\Sigma_{\training}+\frac{1}{2}\gamma I\right)^{-1}T\right].
        \label{eq:D(t)}
\end{equation}
We note that now the limiting value of $D(t\rightarrow\infty)$ is
not zero, but rather $D_\infty=-\frac{1}{2}\gamma\left(\Sigma_{\training}+\frac{1}{2}\gamma I\right)^{-1}T$.
Next, we wish to calculate $\mcL_\training=D(t)^{T}\Sigma_{\training}D(t)$
and $\mathcal{\tilde{L}}_{\gen}=D(t)^{T}\Sigma_{\gen}D(t)$,
where we emphasize that in both cases $D(t)$ is given by Eq. (\ref{eq:D(t)})
and depends on $\Sigma_{\training}$. As described in the main text, it is a good approximation to set $\Sigma_{\gen}$ to be the identity matrix. For
convenience, we will write both cases by $\mcL_{\training/\gen}=D(t)^{T}QD(t),$
where $Q=\Sigma_{\training}$ for the train and $Q=\mI$ (the identity
matrix) for the generalization.

We continue by diagonalizing $\Sigma_{\training}$;
we write $M=P^{T}\Sigma_{ \training }P$, where $M$ is a diagonal
matrix whose eigenvalues follow the MP distribution. Hence, we obtain
\begin{equation}
    \mcL_{\training/\gen}=\bar{D}(t)^{T}\bar{Q}\bar{D}(t),
    \label{eq:loss_after_diagonalization}
\end{equation}
where $\bar{Q}=M,I$ for the train, generalization correspondingly, and $\bar{D}(t)$
is given by
\begin{equation}
    \bar{D}(t)=e^{-2\eta\left(M+\frac{1}{2}\gamma I\right)t}\left[\bar{D}(0)+\frac{1}{2}\gamma\left(M+\frac{1}{2}\gamma I\right)^{-1}\bar{T}\right]-\frac{1}{2}\gamma\left(M+\frac{1}{2}\gamma I\right)^{-1}\bar{T},
\end{equation}
where $\bar{D}(t)=P^{T}D(t),\bar{T}=P^{T}T$. We notice now that the
expression in Eq. (\ref{eq:loss_after_diagonalization}) involves terms in the form of: $V^{T}f(M)W$
where $V,W$ are some vectors, and $f(M)$ is some function of the
diagonal MP matrix. If $V,W$ are random vectors in a large dimension,
we can approximate that
\begin{equation}
    V^{T}f(M)W=\begin{cases}
        0 & V\neq W,\\
        |V|^{2}\int f(u)p(u)du & V=W,
    \end{cases}\label{eq:large_random_vectors_approximation}
\end{equation}
where $|V|$ is the norm of $V$, and $p(u)$ is the probability density
function of the MP distribution. For example, in our case we will
get that $D^{T}(0)f(M)T=-|T|^{2}\int f(u)p(u)du$ (since $D(0)=S(0)-T)$.
All that is left now is to calculate the expression in Eq. (\ref{eq:loss_after_diagonalization})
explicitly, using the approximation of Eq. (\ref{eq:large_random_vectors_approximation}).
Doing this, at last we arrive into
\begin{equation}
    \mcL_{\training/\gen}=
    \din\int\left(|S(0)|^{2}e^{-4\eta\left(u+\frac{1}{2}\gamma\right)t}+|T|^{2}\left(\frac{e^{-2\eta\left(u+\frac{1}{2}\gamma\right)t}u+\frac{1}{2}\gamma}{u+\frac{1}{2}\gamma}\right)^{2}\right)q_\mathrm{\training/gen}p(u)du,
\end{equation}
where $q_\training=u$ and $q_\gen=1$. By also setting the student initialization and teacher vector norms to $|S(0)|, |T| \simeq 1/\sqrt{2\din}$ (as done in the main text), we finally get
\begin{equation}
    \mcL_{\mathrm{\training/gen}}=
    \frac{1}{2}
    \int\left(e^{-4\eta\left(u+\frac{1}{2}\gamma\right)t}
    +\left(\frac{e^{-2\eta\left(u+\frac{1}{2}\gamma\right)t}u+\frac{1}{2}\gamma}{u+\frac{1}{2}\gamma}\right)^{2}\right)q_\mathrm{\training/gen}p(u)du.
\end{equation}


\section{Derivation for the 2-layer network}
\label{sec:app_2layer}

\alonb{TODO: Check all of the $din \times dout$ and etc.}Here, we provide supplementary details on the derivation of \cref{eq:2_layer_loss}.
We consider the addition of a single hidden linear layer, where the teacher network function is $f(x)= (T^{(1)})^T (T^{(0)})^T  x$, where $T^{(0)} \in \mathbb{R}^{ \din\times d_h}$, $T^{(1)} \in \mathbb{R}^{ d_h \times \dout}$ and $d_h$ is the width of the hidden layer. Similarly, the student network is defined by two matrices $S^{(0)}, S^{(1)}$. The empirical training loss over a sample set $\{x_i\}_{i=1}^N$ reads
\begin{align}
    \mathcal{L}_\training 
    &= 
    \frac{1}{N_\training \dout}
    \sum_{i=1}^{N_\training }
    \left( 
    (S^{(1)} )^T (S^{(0)} )^T  x_i 
    -( T^{(1)} ) )^T (T^{(0)} ) )^Tx_i
    \right)^2\ .
\end{align}
In this setup the weights are drawn at initialization from normal distributions $S^{(0)}_0, T^{(0)} \sim \mathcal{N}(0,1/({2 \din }d_{h}))$ , $S_0^{(1)}, T^{(1)} \sim \mathcal{N}(0,1/({2 \dout }d_{h}))$. 
Next, we define $T= T^{(0)} T^{(1)} \in \mathbb{R}^{\din \times \dout }$ and derive the gradient flow equations for the system 
\begin{align}
    \dot{S}_t^{(0)}
    =
    -\frac{2\eta_0}{\dout} 
     \Sigma_\training  
     \left( 
    S^{(0)}_t S^{(1)}_t   
    - T
    \right) 
    (S^{(1)}_t)^T
    ,
    \quad
     \dot{S}_t^{(1)}
    =
    -\frac{2\eta_0}{\dout} 
    (S^{(0)}_t ) ^T
     \Sigma_\training 
     \left( 
      S^{(0)}_t S^{(1)}_t
    - T
    \right).
\end{align}
defining $D_t= S_t^{(0)}S_t^{(1)}  - T$, and noting that 
$\dot{D}_t=  S_t^{(0)}\dot{S}_t^{(1)} +  \dot{S}_t^{(0)}S_t^{(1)}$, we arrive at the equations quoted in the main text
\begin{align}
    \dot{D}_t
    =
    -
    2\eta_0 \frac{h}{d_\mathrm{out}^2}  \Sigma_\training  D 
    ,
    \qquad
    \dot{h}_t
    =
    -8\eta_0 (T+D)^T \Sigma_\training D.
\end{align}
Here, $h=Tr[H]/2=\|S^{(0)}\|^2/2 + \|S^{(1)}\|^2/2$, where $H=\gradtheta^T \gradtheta \mathcal{L}_\training$ is the Hessian matrix and $\theta\equiv\{S^{(0)},S^{(1)}\}$.
Although \cref{eq:GF_2layer} describes a set of coupled equations, we note that the solution for $h_t$ can be simplified when considering

the limit of small $\eta_0 \ll1$, as we may ignore the time evolution and consider the trace (or kernel) as fixed to its initialization value, which is $h_0\simeq 1/2$ for $d_h \gg d_\mathrm{out}$. In that case the loss solutions are a simple modification to the ones given in the previous sections, with the replacement $\eta_0 \to \eta_0 /(2 d_\mathrm{out}^2)$.
Subsequently, the training/generalization performance metrics are 
\begin{align}
  \mathcal{L}_\mathrm{\training/gen}^\mathrm{2-layer} 
  &= 
  \|D_0 \|^2
  \mathcal{L}_\mathrm{\training/gen}^\mathrm{1-layer}
  \left(
  \frac{\eta_0 }{2 \dout^2}, \lambda, t
  \right),
  \quad
    \mathcal{A}_\mathrm{\training/gen} = 
  1-\frac{\Gamma \left(\frac{\dout}{2},\frac{\dout \epsilon }{2 \mathcal{L}_\mathrm{\training/gen}}\right)}{\Gamma \left(\frac{\dout}{2}\right)} .
\end{align}

\end{document}